\documentclass[sigconf]{acmart}

\usepackage{enumitem}

\AtBeginDocument{%
  \providecommand\BibTeX{{%
    \normalfont B\kern-0.5em{\scshape i\kern-0.25em b}\kern-0.8em\TeX}}}

\copyrightyear{2022}
\acmYear{2022}
\setcopyright{acmlicensed}\acmConference[MM '22]{Proceedings of the 30th ACM International Conference on Multimedia}{October 10--14, 2022}{Lisboa, Portugal}
\acmBooktitle{Proceedings of the 30th ACM International Conference on Multimedia (MM '22), October 10--14, 2022, Lisboa, Portugal} \acmPrice{15.00}
\acmDOI{10.1145/3503161.3548081} \acmISBN{978-1-4503-9203-7/22/10}

\begin{document}


\title{Lip-to-Speech Synthesis for Arbitrary Speakers in the Wild}

\author{Sindhu B Hegde}
\authornote{All three authors contributed equally to this research.}
\email{sindhu.hegde@research.iiit.ac.in}
\affiliation{%
  \institution{International Institute of Information} \country{Technology, Hyderabad, India}
}

\author{K R Prajwal}
\authornote{Work done at IIIT Hyderabad}
\email{prajwal@robots.ox.ac.uk}
\authornotemark[1]
\affiliation{%
  \institution{University of Oxford}
  \country{United Kingdom}
}

\author{Rudrabha Mukhopadhyay}
\email{radrabha.m@research.iiit.ac.in}
\authornotemark[1]
\affiliation{%
  \institution{International Institute of Information} \country{Technology, Hyderabad, India}
}

\author{Vinay P Namboodiri}
\email{vpn22@bath.ac.uk}
\affiliation{%
  \institution{University of Bath} 
  \country{United Kingdom}
}

\author{C. V. Jawahar}
\email{jawahar@iiit.ac.in}
\affiliation{%
  \institution{International Institute of Information} \country{Technology, Hyderabad, India}
}

\renewcommand{\shortauthors}{Sindhu B Hegde et al.}

\begin{abstract}
  In this work, we address the problem of generating speech from silent lip videos for any speaker in the wild. In stark contrast to previous works, our method (i) is not restricted to a fixed number of speakers, (ii) does not explicitly impose constraints on the domain or the vocabulary and (iii) deals with videos that are recorded in the wild as opposed to within laboratory settings. The task presents a host of challenges, with the key one being that many features of the desired target speech, like voice, pitch and linguistic content, cannot be entirely inferred from the silent face video. In order to handle these stochastic variations, we propose a new VAE-GAN architecture that learns to associate the lip and speech sequences amidst the variations. With the help of multiple powerful discriminators that guide the training process, our generator learns to synthesize speech sequences in any voice for the lip movements of any person. Extensive experiments on multiple datasets show that we outperform all baselines by a large margin. Further, our network can be fine-tuned on videos of specific identities to achieve a performance comparable to single-speaker models that are trained on $4\times$ more data. We conduct numerous ablation studies to analyze the effect of different modules of our architecture. We also provide a demo video that demonstrates several qualitative results along with the code and trained models on our website\footnote{\url{http://cvit.iiit.ac.in/research/projects/cvit-projects/lip-to-speech-synthesis}}.
\end{abstract}

\ccsdesc[500]{Computing methodologies~Reconstruction}
\ccsdesc[500]{Computing methodologies~Neural networks}

\keywords{lip-to-speech, speech synthesis, hybrid vae-gan, talking-face videos}

\vspace{-20pt}
\begin{teaserfigure}
  \includegraphics[width=\textwidth]{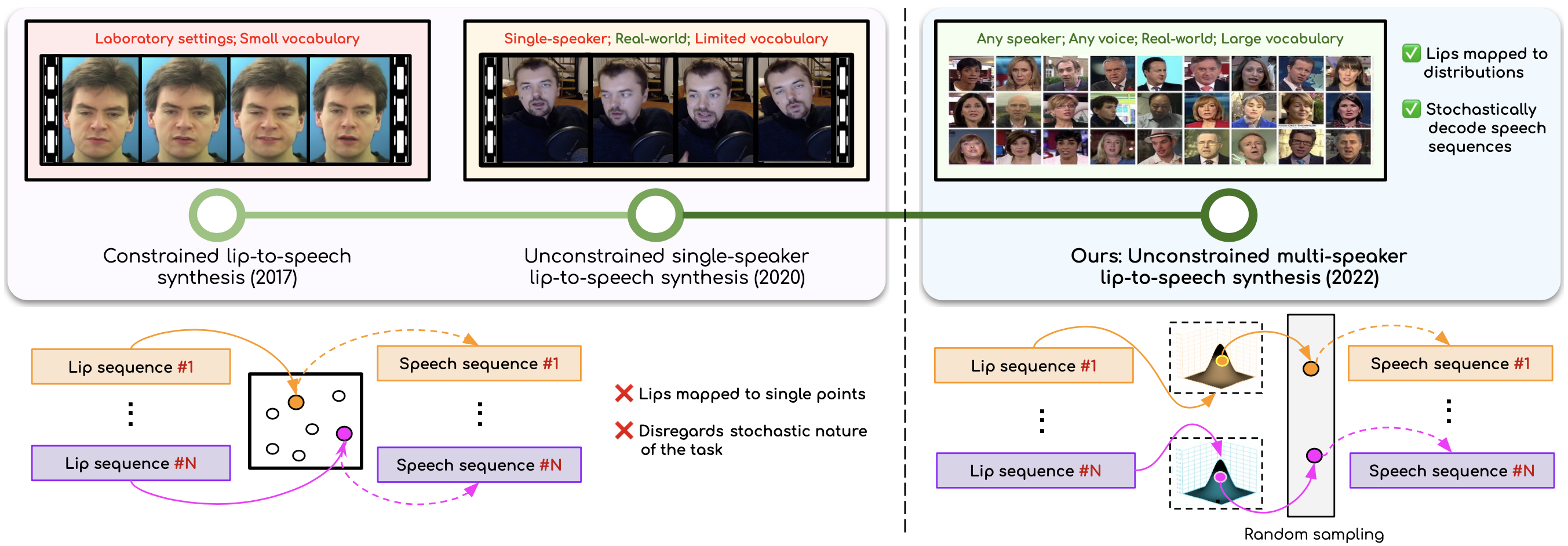}
  \vspace{-20pt}
  \caption{We address the problem of generating speech from silent lip videos for any speaker in the wild. Previous works train either on large amounts of data of isolated speakers or in laboratory settings with a limited vocabulary. Conversely, we can generate speech for the lip movements of arbitrary identities in any voice without additional speaker-specific fine-tuning. Our new VAE-GAN approach allows us to learn strong audio-visual associations despite the ambiguous nature of the task.}
  \label{fig:teaser}
\end{teaserfigure} 

\maketitle
\section{Introduction}
\label{section:introduction}

As the world's communication becomes increasingly digital, it is also becoming increasingly visual. From video calls to movies to YouTube videos, there is a surge in video content consumption. Naturally, understanding and enabling applications for talking-face videos~\cite{wav2lip,Prajwal_2020_CVPR,chung2017lip,conversation,cocktail_party,Afouras2018DeepLR} has been an active area of research in recent years. Tasks such as speech/text-based lip synthesis~\cite{Kumar2018ObamaNetPL, wav2lip, lipgan} have witnessed tremendous advancements. The opposites of these tasks, namely, lip-to-text generation and lip-to-speech generation, both falling under the umbrella of ``lip-reading'', have proven far more challenging. For the task of lip-to-text generation, multiple impressive works have pushed the boundaries with models that work for any speaker in the wild. However, its sibling task, lip-to-speech synthesis, has not yet witnessed a similar advancement in such unconstrained settings.\\

\noindent
\textbf{Lip-to-Speech Synthesis for Arbitrary Identities:}
The goal of lip-to-speech synthesis is to generate meaningful speech for a silent talking-face video. Previous works in this space have focused on training models that work for a fixed set of speakers. They achieve impressive results but rely on videos recorded in laboratory settings~\cite{cooke2006audio,harte2015tcd} or require tens of hours of single-speaker data~\cite{Prajwal_2020_CVPR} when working with real-world videos. This makes the previous methods hard to scale to the large number of identities in the wild. 

Our goal in this work is to perform lip-to-speech synthesis for silent videos of any identity. This allows us to produce results on any speaker at test time. We also show that we can further fine-tune on videos of a single speaker, if necessary, and achieve similar performance to single-speaker models but with $4\times$ lesser data. \\

\noindent
\textbf{Overarching Challenges:}
The set of challenges in our task can be divided into two major groups: (i) challenges of lip-to-speech generation and (ii) challenges in handling the large variations in identities. In lip-to-speech generation, deciphering the uttered words is ambiguous, e.g. lip movements of ``\underline{p}at'', ``\underline{b}at'' and ``\underline{m}at'' are the same but map to different speech outputs. The second set of challenges is unique to the task in this work and was not faced by previous single-speaker models. The diversity in voices, accents and speaking styles makes it difficult to learn the lip-speech correspondences. Generating continuous speech data is also far more challenging when there are no constraints on identity and voice. Finally, more speakers usually mean more variations in terms of content and topics spoken. Given this second set of challenges, we must ask ourselves: \textit{Can single-speaker methods be directly extended for unconstrained, real-world multi-speaker lip-to-speech synthesis?}\\

\noindent
\textbf{Overview of this Work:}
The key idea of this work is to allow the model to handle the highly stochastic nature of the task. Unlike the constrained single-speaker case, the model has limited knowledge about the topic being spoken. It needs to determine which voice to generate in, the pitch, accent, emotion, prosody and tone. All these aspects vary stochastically due to inadequate priors in the input. Existing single-speaker models trained using an L$_1$ reconstruction loss enforces a highly constrained one-to-one correspondence between the input and the output, which, as we will see, is detrimental to this task.  

In this work, we propose a novel VAE-GAN architecture whose core idea is to map the input lip sequence to an output distribution of plausible speech sequences. We are the first to handle the issue of ambiguities in lip-to-speech synthesis explicitly. Through extensive quantitative and qualitative comparisons, we show that this is very helpful in such an unconstrained setting. In addition to the variational architecture, we add a variety of perceptual loss functions to ensure the realism and style of the generated speech. We show that these discriminators (GAN discriminator to enforce the generation quality and voice discriminator to enforce the voice quality) play an essential role when learning such an unconstrained task. Our model not just handles videos of arbitrary speakers but also makes the single-speaker lip-to-speech synthesis task much more scalable. We show that we can match the quality of the current state-of-the-art single-speaker models while using $4\times$ lesser training data.
Our key contributions/claims in this work are:

\begin{itemize}[topsep=0.8pt, noitemsep, leftmargin=4mm]

    \item We address the problem of lip-to-speech synthesis in the wild, with no explicit constraints on the number of speakers and vocabulary. This allows us to, for the first time, generate speech for any person's silent lip movements in any voice.  
    
    \item We distill the content information from speech sequences and align them with the corresponding lip movements using our novel VAE-GAN architecture.
    
    \item Our pre-trained model can be further fine-tuned and personalized to specific speakers in a data-efficient manner compared to the current single-speaker models trained from scratch. We show that our network achieves comparable performance while using only $25\%$ of the training data.

\end{itemize}

\section{Related Work}
\label{section:related_work}

\subsection{Constrained Multi-speaker Lip-to-speech}
The problem of lip-to-speech synthesis has been receiving growing attention in recent times. One of the initial works~\cite{ephrat2017vid2speech} in constrained laboratory settings used a 2D convolution-based encoder-decoder architecture to learn a mapping between lip movements and LPC features of the corresponding speech. The network is trained on the GRID corpus~\cite{cooke2006audio}, containing lab-recorded videos from $34$ speakers. There have been follow-up works~\cite{Ephrat2017ImprovedSR, Akbari2017Lip2AudspecSR, l2sacmmm, kumar2019lipper, vougioukas2019video} that train in similar settings~\cite{cooke2006audio,harte2015tcd}, containing speakers with limited head movements and a small vocabulary. However, we observe that the performance of these works severely degrades when directly extended to unconstrained settings comprising in-the-wild videos with large variations in vocabulary, speakers and head movements.

\subsection{Single-speaker Lip-to-speech}
In order to perform lip-to-speech synthesis for in the wild silent videos, a more recent work, Lip2Wav~\cite{Prajwal_2020_CVPR} proposes the idea of learning a model by training on large amounts of single-speaker data. Lip2Wav demonstrates impressive results in real-world settings by utilizing $\approx 20$ hours of data for isolated speakers. The sheer amount of data per speaker allows the model to learn fine-grained speaker-specific attributes. The same work also shows preliminary results on word-level multi-speaker lip-to-speech using LRW~\cite{chung2016lip} dataset. In Table~\ref{tab:work_comparison}, we contrast our task against the previous works. Most of the earlier works function under one or more constraints. As we will show later, these methods do not scale to the case of ``generating speech for any identity in any voice''. We discuss the reasons and describe how our novel approach addresses these issues. 

\begin{table}[h]
\centering
  \small
  \setlength{\tabcolsep}{2pt}
 
  \caption{Major differences between our approach and the existing approaches. Our work deals with the most challenging task in this space.}
  
  \vspace{-10pt}
  \begin{tabular}{c|c|c|c|c}
  
    \hline
    \textbf{Approach} & \textbf{vocab.} & \textbf{natural} & \textbf{training data per} & \textbf{zero-shot gen.} \\
    & \textbf{size} & \textbf{setting?} & \textbf{spkr (in mins.)} & \textbf{(unseen spkrs.)}\\

    \hline

    Vid2Speech~\cite{ephrat2017vid2speech} & 56 & $\times$ & 48 &$\times$\\
    
    Ephrat et.al~\cite{Ephrat2017ImprovedSR} & 82  & $\times$ & 30 & $\times$\\
    
    GAN based~\cite{vougioukas2019video} & 82  & $\times$ & 30 &$\times$\\
    
    Lip2AudSpec~\cite{Akbari2017Lip2AudspecSR} & 56 & $\times$ & 48 & $\times$\\
    
    Lip2Wav~\cite{Prajwal_2020_CVPR} & $\approx5K$  & \checkmark & 1200 & $\times$\\
    
    \hline
    
    \textbf{Ours} & \textbf{50K+} & \textbf{\checkmark} & \textbf{3} & \textbf{\checkmark}\\
    
    \hline
  
  \end{tabular}
  \label{tab:work_comparison}
  \vspace{-10pt}
\end{table}

\subsection{Lip-to-text Generation}
\label{related_work:lip_to_text}
A closely related task to lip-to-speech generation is lip-to-text generation, usually referred to as ``lip reading''. Early works focused on obtaining single-word labels by posing it as a classification problem~\cite{chung2016lip, SynthesizingObama_tog_2017}. More recent works can produce sentence-level predictions using different models with losses like CTC~\cite{LipNet_arxiv_2016, LCANet_fg_2018} and models ranging from  LSTMs~\cite{chung2017lip} to Transformers~\cite{DeepAVLR_tpami_2018, SubWordLR_arxiv_2022}.   

Synthesizing speech from lips is far more challenging than generating text from lips due to the following reasons: (i) Lip-to-text models only need to transcribe the content (words), whereas in lip-to-speech, along with the content, other speaker attributes like voice and prosody also need to be modelled; (ii) Lip-to-speech deals with continuous outputs (harder for learning), whereas lip-to-text has the luxury of generating discrete tokens. Thus, although there has been tremendous progress in unconstrained lip-to-text generation, lip-to-speech generation in unconstrained settings, which is the focus of our work, still has a large room for improvement.

\section{VAE-GAN Architecture}
\label{section:methodology}

\begin{figure*}[t]
  \includegraphics[width=\textwidth]{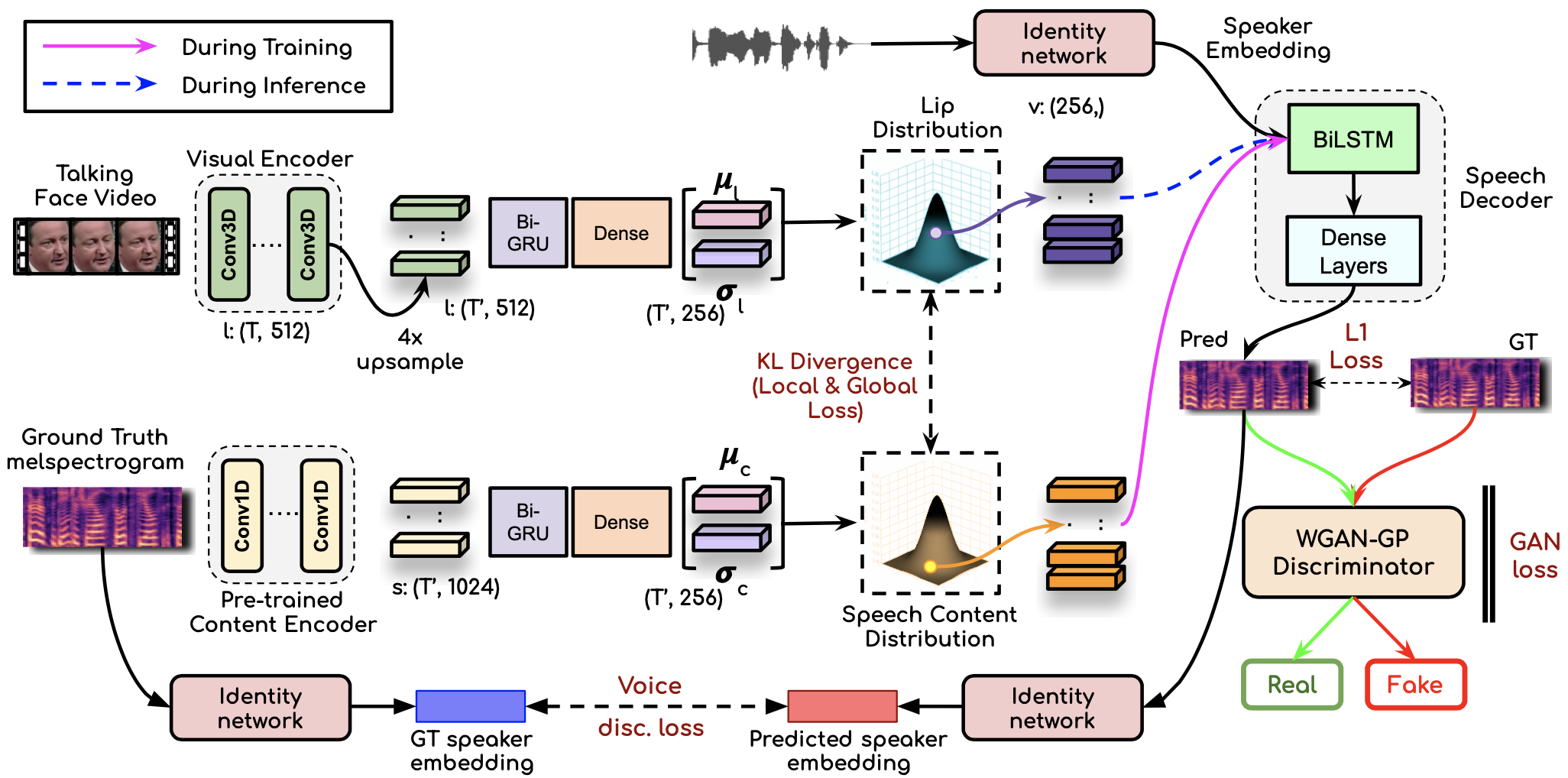}
  \vspace{-10pt}
  \caption{We propose a novel VAE-GAN architecture for our task. While previous approaches enforce a one-to-one mapping between lip and speech sequences, we deal with the task's ambiguities differently. We map the speech content (ASR representations) and the lip sequence to similar distributions and use a decoder to generate realistic speech outputs from this latent space. Additional discriminators enable high-fidelity generation in such unconstrained settings.}
  \label{fig:architecture}
  \vspace{-10pt}
\end{figure*}

Given a sequence of lip movements $L = (L_{1}, L_{2}, ..., L_{T})$ and a speaker identity vector $V$, the goal is to generate speech segment $S = (S_{1}, S_{2}, ..., S_{T'})$ corresponding to the lip movements $L$ and in the voice of $V$. We start our discussion by examining the issues in previous methods and propose appropriate changes to enable learning in a significantly more unconstrained multi-speaker setting. 

\subsection{Fundamental issues in Previous Works}
\subsubsection{Stochastic Nature in Lip-to-Speech Synthesis}
\label{subsub:stochastic}
All of the previous works aim to map the input lip sequence to a single speech sequence, i.e., they do not account for stochastic nature of the task. The stochasticity arises due to inadequate priors, i.e., the speech cannot be entirely inferred from the lip movements due to the homophene ambiguity. But additional ambiguities are introduced as we move from laboratory settings to utterances in real-world videos~\cite{Prajwal_2020_CVPR}, where even the single-speaker case becomes challenging when the speech is ``freely uttered'': it can have varying decibel levels (no concrete correlation to lips), stress on particular phonemes, and even transient lip motion during pauses. 

\subsubsection{Scaling to Multi-speaker Lip-to-Speech}
As we move further into the multi-speaker case, the task becomes severely ill-posed and extremely challenging. Given only the lip movements and a voice token, there are many stochastically varying factors that cannot be inferred from either of the inputs. In addition to the ambiguities mentioned in Section~\ref{subsub:stochastic}, each speaker can have distinct speaking styles and lip shapes in the multi-speaker setting. The large variation in voices and accents also influences how the phonemes are uttered. Such variations cannot be adequately captured in the voice token input. As none of the existing models handle these issues, even in the single-speaker case, they do not scale well to multi-speaker lip-to-speech. 

\subsubsection{What ``Space'' is Right to Learn these Ambiguous Audio-Visual Correspondences?}
Finally, the existing models struggle to learn in the unconstrained multi-speaker scenario because their learning signal comes from the level of raw spectrograms. This is because most of the current models use a \textit{visual encoder - speech decoder} approach with only an L$_1$ reconstruction loss. Given a large amount of stochastic variations in both the visual and speech modalities, we argue that it is beneficial to learn speech-lip correspondences in the feature space, whose benefits have been well-studied in the literature~\cite{oord2018representation,chen2020simple,percept1,percept2,percept3}. The intuition is that the low-level variations are more meaningfully represented in the feature space. For example, matching the lip shapes of ``ma'' with its instances in the speech in different voices will be far easier in a feature space that is voice invariant and contains only the content information from the speech sequences. We build upon this intuition to arrive at our core idea. 

\subsection{Our Core Idea}
Our core idea is two-fold. Firstly, we want to match the distributions (Figure~\ref{fig:teaser}) of (i) the lip sequence and (ii) the content from the speech sequence in the latent feature space to allow the model to handle the stochastic variations mentioned above. Secondly, we want to learn a decoder that decodes meaningful speech samples from this latent space while also conditioning on a speaker identity embedding that provides the voice information.

Concretely, we first represent each input lip sequence as a distribution (instead of a single vector) and match it to the corresponding speech content distribution. The intuition for matching at the level of distributions is that it allows the ambiguities to be meaningfully represented by allowing a ``one-to-many'' correspondence. Once we have such a shared latent space, the second step is to decode samples from this latent distribution and generate meaningful speech. We realize these ideas in the following manner. We use a standard automatic speech recognition (ASR) model to extract content information from the input speech sequences. A variational auto-encoder~\cite{vae} then maps the speech content information to a shared latent space and decodes the samples from this latent space to real speech sequences. We have an additional visual encoder that maps the lip sequences to the same shared latent space. We tie these two latent distributions together using the Kullback-Leibler Divergence (KL) loss~\cite{kl_div} as illustrated in Figure~\ref{fig:architecture}. Finally, we sample points from these distributions and feed them to a speech decoder along with the speaker identity embedding to generate intelligible speech sequences. We delve into each of the modules below.

\subsubsection{Visual Encoder}
We adopt the visual encoder used in several previous models that aim to learn audio-visual correspondence~\cite{avobjects, conversation, Chung16a, perfect_match}. Our visual encoder consists of 3D convolutional layers, with only the first layer having a temporal receptive field of $5$ frames. It provides a good trade-off between speed and capturing short-range temporal information. The visual encoder inputs a spatio-temporal volume $L: (T, 96, 96, 3)$ and outputs 1D embeddings $l: (T, 512)$ at each time-step. We perform $4\times$ temporal upsampling using nearest-neighbor interpolation on $l$ to match the speech time-steps $T'$.

\subsubsection{Speech Content Encoder}
We believe that lip movements primarily represent the content information present in a speech sequence. Thus, before matching with the lip distribution, we need to distill the content from speech segment. We achieve this using a standard pre-trained ASR network~\cite{deep_speech_2}. The melspectrogram is passed through this frozen encoder to generate a $T'\times1024$ dimensional embedding denoted by $c$. Thus, we separate the voice information from the speech representations, which, as we will see later, is crucial to our training strategy. 

\subsubsection{Variational Auto Encoder Based Approach \& Latent Distribution Matching}
We now map both, lip and speech content embeddings, $l$ and $c$ to Gaussian distributions with a diagonal covariance matrix: $\mathcal{N}(\mu_l, \sigma_l)$ and $\mathcal{N}(\mu_c, \sigma_c)$ , where $(\mu_l, \sigma_l), (\mu_c, \sigma_c)$ are obtained using two projection modules $P_l$ and $P_c$. Both these modules contain a bi-directional GRU~\cite{gru} followed by ReLU-activated fully-connected layer. The bi-directional GRU helps to capture contextual information in both directions at each time-step. Now that we have two distributions, one corresponding to the speech content, another corresponding to the lips, random points $c_p$ and $l_p$ are sampled from these distributions using the re-parametrization trick~\cite{vae}. We can clearly see that we no longer have a ``single value'' for each input lip or speech sequence, but rather two probability distributions for these inputs. 

Our final step is to tie these distributions together, i.e., we want the lip distribution $\mathcal{N}(\mu_l, \sigma_l)$ to be close to the speech content distribution $\mathcal{N}(\mu_c, \sigma_c)$. If we do this, then we can train a decoder that decodes speech samples from the content distribution $\mathcal{N}(\mu_c, \sigma_c)$ and also use it to decode from points in the lip distribution. Thus, we minimize the Kullback-Leibler Divergence (KL) loss~\cite{kl_div} between these two distributions:  

\vspace{-10pt}
\begin{equation}
    L_\mathrm{kl_{global}} = \frac{1}{N} \sum^{N}_{i=1} KL[\mathcal{N}(\mu_c, \sigma_c) || \mathcal{N}(\mu_{l}, \sigma_{l})]
\label{eq:klglobal}
\end{equation}

\begin{figure}[t]
  \includegraphics[width=\linewidth]{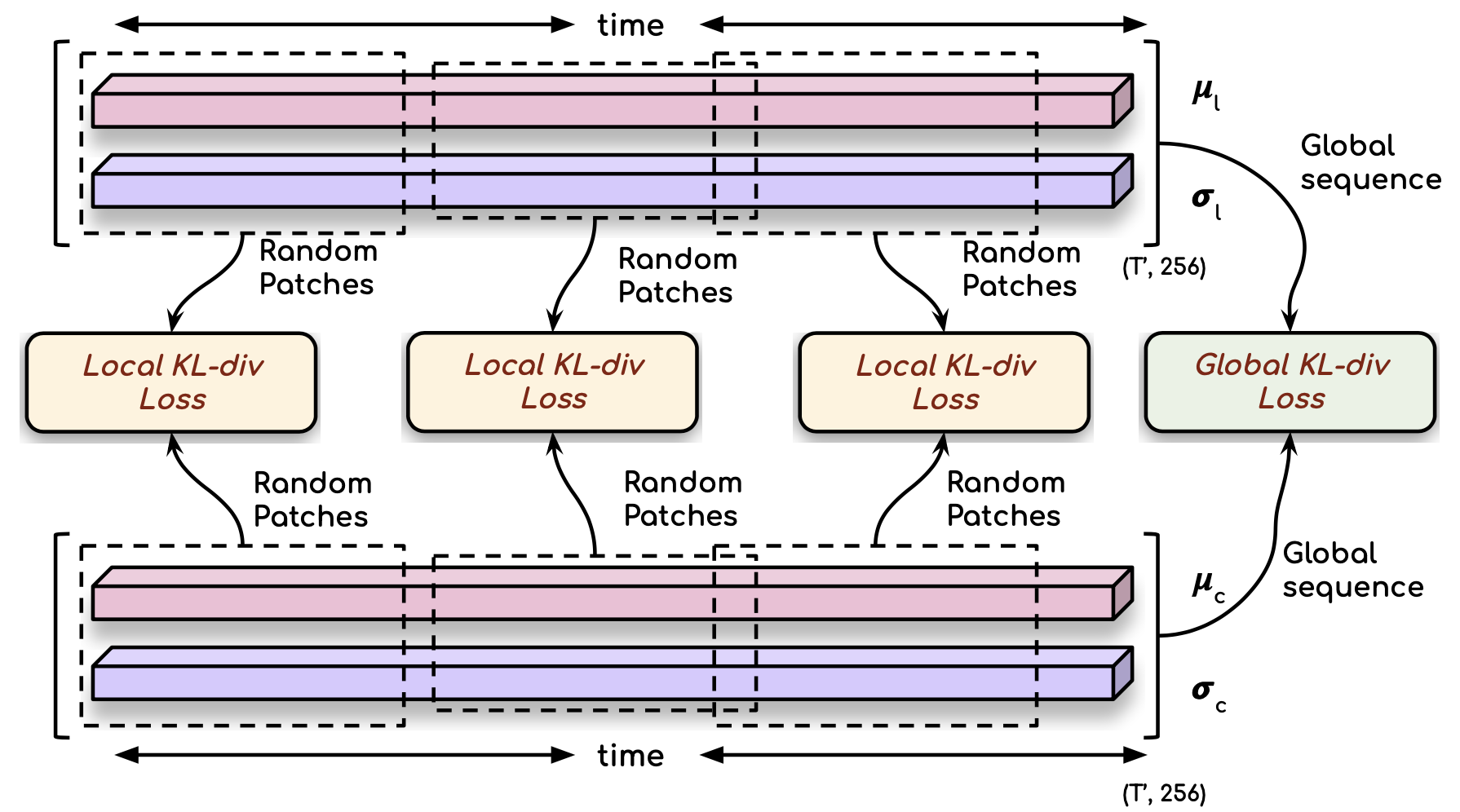}
  \vspace{-20pt}
  \caption{In addition to using a global KL-divergence loss to tie the lip and speech content distributions, we also enforce these distributions to be temporally aligned by minimizing a local KL-divergence loss on random smaller time segments. The intuition is that lips and speech are locally aligned in time, in the form of visemes and phonemes.}
  \label{fig:local_kl_div}
  \vspace{-20pt}
\end{figure}

We term $L_{kl_{global}}$ as the global KL-divergence loss since the distributions are created by considering the complete sequence of speech and lip movements. To further improve the alignment, inspired from~\cite{Demir_2018_PatchGAN}, we take random corresponding temporal segments of the distributions and align them by minimizing a ``local'' KL-divergence loss (Figure~\ref{fig:local_kl_div}). We choose $R=10$ small temporal segments for each batch sample and use its $(\mu^r, \sigma^r)$ to minimize Equation~\ref{eq:kllocal}: 

\vspace{-10pt}
\begin{equation}
    L_\mathrm{kl_{local}} = \frac{1}{N} \sum^{N}_{i=1}\sum^{R}_{r=1} KL[\mathcal{N}(\mu^r_c, \sigma^r_c) || \mathcal{N}(\mu^r_{l}, \sigma^r_{l})]
\label{eq:kllocal}
\end{equation}

Here, $\mathcal{N}(\mu^r, \sigma^r)$ are $r^{th}$ random patches sampled from the lip and content distributions along the temporal dimension. Binding the distributions at the local level is crucial as phoneme-viseme mappings occur locally rather than globally. We show the importance of employing both local and global KL-divergence losses in Table~\ref{table:global_vs_local}. Since the two distributions are aligned using the KL-divergence loss, we can sample from the lip distribution during inference while sampling from the speech content one during training. 

\subsubsection{Speaker Embedding}
\label{subsec:speaker_emb}
While the visual/content encoder specifies ``what to utter'', we also need an input for ``which voice to utter in''. While this can be done by representing each speaker in the dataset with a one-hot vector, it does not generalize to new speakers during inference. Instead, we adopt a recent advancement~\cite{sv2tts} in training multi-speaker text-to-speech models, where a pre-trained identity network\footnote{\url{github.com/CorentinJ/Real-Time-Voice-Cloning}} containing the embedding with voice information is used. The speaker embedding can be obtained for any voice, given just one second of the voice sample. For each video in the dataset, we generate a $256$-dimensional speaker embedding using a random one-second segment of the audio. We apply a ReLU-activated fully-connected layer to this pre-trained speaker embedding input $V$ before feeding it to the decoder.

\subsubsection{Speech Decoder}
Our final step is to train a module that can generate speech segments given points sampled from the above created joint latent space. It is clear that we need to feed points from the lip distribution at test time, as we do not have the speech. During training, we have three ways of sampling the points: (i) only from the lip distribution, (ii) only from the speech content distribution and (iii) alternately sample from both the distributions. We hypothesize that learning with points from (ii) is far easier and allows the network to learn excellent latent representations of the speech content. As the distributions are being matched in the latent space, learning accurate, meaningful representations of one of them can be quite beneficial for learning the joint space. Indeed, we observed that good convergence and intelligible speech both at training and test time could be achieved only by training the decoder on points sampled from the speech content distribution. The points are sampled using the re-parametrization trick~\cite{vae} and are of the shape $(T', 256)$. Along with the sampled points from content ($c_p$) or lip distribution ($l_p$), the decoder ingests the speaker embedding input $V$, which is concatenated with the sampled points. The concatenated content-voice feature vectors $[(c_p | l_p); V]: (T', 512)$ contain information on both ``what to utter'' and ``the voice to utter in''. We use a bi-directional LSTM layer followed by $4$ dense layers to decode the melspectrogram segment $(T', 80)$ from the concatenated feature vector. During training, the generator $G$ ingests speech content $c$ and speaker embedding $V$ and minimizes the L$_1$ reconstruction loss between the generated speech and the ground-truth speech $S$:

\vspace{-10pt}   
\begin{equation}
L_{\mathrm{r}} = \frac{1}{N} \sum^{N}_{i=1} ||G(c, V) - S||_1
\label{eqn:recon}
\end{equation}

Note that during training, the generator network is essentially a VAE for the speech with an additional KL loss constraint on its latent space. Because of the KL loss, we can sample from the speech distribution during training, and during inference, when the speech is absent, we can sample and decode points from the lip distribution. Therefore, during inference, we predict $G(l, V)$ as our output. Since we employ a content encoder during training, the decoder is forced to condition on the speaker embedding for the voice information. The content encoder distills only the content information from a speech sequence and does not leak the voice information, allowing us to maintain good voice quality even at test time when we decode from the lip distribution. We now describe additional discriminators that we will use along with our generator to improve the quality and accuracy of the generated speech outputs.

\subsubsection{Enforcing Realism with a VAE + GAN}
In our experiments, we observed that generating realistic samples for such a diverse set of voices, accents and speaking styles, using a plain L$_1$ reconstruction loss produced unrealistic, unintelligible samples (Table~\ref{table:discs}). We hypothesize that this occurs because of the known issue of the L$_1$ loss regressing to the mean. Multiple works in the past~\cite{larsen2016autoencoding,rosca2017variational} also point to the benefits of using a GAN along with a VAE. We found that it is highly beneficial to train a WGAN-GP~\cite{wgan} critic in a GAN setup along with our VAE architecture. The critic consists of a series of 1D convolutional layers that takes an audio spectrogram segment of shape $T' \times 80$ as input and outputs a single number as the score. The generator $G$ and the critic $D$ optimize the Wasserstein objective~\cite{wgan_original} along with the gradient penalty~\cite{wgan} in the equations below, where $\hat{S}$ contains all linear interpolates between $S$ and $G(L, V)$:

\vspace{-10pt}
\begin{gather}
    L_\mathrm{adv} = \mathbb{E}_{x\sim S}[D(x)] - \mathbb{E}_{x' \sim G(L, V)}[D(x')]\\
    L_\mathrm{gp} = \mathbb{E}_{\hat{x} \sim \hat{S}}[(\lVert \nabla_{\hat{x}}D(\hat{x}) \rVert - 1)^{2}
\label{eq:wgan}
\end{gather}

\subsubsection{Improving Voice of the Generated Speech}
To ensure that the model learns the voice and other style attributes, we use our pre-trained identity network as described in Section~\ref{subsec:speaker_emb} to penalize the generated speech segments if they do not match the voice/style attributes of the ground-truth speech segment. We train the discriminator network to maximize the cosine similarity $L_{voice}$ between the embeddings of the generated ($V_{gen}$) and the ground-truth ($V_{GT}$) speech segments.

\vspace{-10pt}
\begin{equation}
    L_\mathrm{voice} = \frac{1}{N} \sum^{N}_{i=1}  \dfrac{V_{gen} \cdot V_{GT}}{max(\lVert V_{gen} \rVert_{2} \cdot \lVert V_{GT} \rVert_{2}, \epsilon)}
\label{eq:voice_loss}
\end{equation}

\subsection{Training Settings \& Inference}
The complete loss function to train our network is the weighted summation of all the above losses:

\vspace{-15pt}
\begin{multline}
    L_\mathrm{G} = \lambda_{r}L_\mathrm{r} + \lambda_{k_{global}}L_\mathrm{KL_{global}} + \lambda_{k_{local}}L_\mathrm{KL_{local}} \\+ \lambda_\mathrm{voice}L_{voice} + \min_{G}\max_{D}L_\mathrm{adv} + \lambda_\mathrm{g}L_\mathrm{gp}
\end{multline}

In our experiments we set $\lambda_{r}=10$, $\lambda_{k_{global}}=5$, $\lambda_{k_{local}}=5$ and $\lambda_{voice}=5$. We follow the pre-processing procedures of Lip2Wav~\cite{Prajwal_2020_CVPR} to detect and extract face crops from training videos. We create video inputs by randomly sampling a window of $T=25$ ($1$ sec.) contiguous face crops resized to $96\times96$. The corresponding audio segment is sampled at 16kHz. We compute STFT with a hop length of $10ms$ and a window length of $25ms$. We finally obtain melspectrograms with $80$ mel-bands and $T'=100$ mel time-steps ($1$ sec.). We use a batch size of $32$ and RMSProp~\cite{rmsprop} optimizer with an initial learning rate of $0.00005$ for both the generator and the discriminator, which is advised for training a WGAN model~\cite{wgan}. The generator is trained every five discriminator iterations following~\cite{wgan}. As this is a WGAN, the discriminator loss shows the progress of training and correlates with the quality of the generated samples. Hence, we stop the training once the discriminator loss does not improve for $10$ epochs. During inference, we feed the speaker embedding and the lip distribution to the decoder instead of the content distribution. Since our model can take a variable number of time steps as input, it can directly generate for any length of video without any further changes. 

\vspace{-5pt}
\subsubsection{Datasets and Training Strategy}
Our primary focus is to synthesize speech for silent lip videos in unconstrained settings; we intend to make our model identity-agnostic and work for a larger vocabulary. But, for the sake of comparison with previous works, we also train our model on the lab-recorded constrained GRID~\cite{cooke2006audio} and TCD-TIMIT~\cite{harte2015tcd} datasets. We use the speaker-independent train-test setting as~\cite{vougioukas2019video} for GRID and
single-speaker lip-to-speech setting as~\cite{Prajwal_2020_CVPR} for TCD-TIMIT dataset. For unconstrained evaluation, we first train our model on the word-level LRW data~\cite{chung2016lip}. Next, we use the complete LRS2 dataset~\cite{chung2017lip} (both train and pre-train sets), which contains sentences and phrases as opposed to specific words. The LRS2 data comprises thousands of speakers from BBC programs with a vocabulary of $59k$ and $2M$ word instances. A large number of speakers and vast vocabulary covered in both of these datasets encourage our model to be speaker agnostic and pose no limitations on the vocabulary size. 
    
\section{Experiments}
\label{section:experiments}

We evaluate our model against various baseline methods in two settings: (i) laboratory setting videos and (ii) in the wild videos.

\subsection{Evaluation in Constrained Settings}

\subsubsection{Baselines}
We compare our model with the following existing lip-to-speech methods: (i) Improved Vid2Speech~\cite{Ephrat2017ImprovedSR}, (ii) GAN-based~\cite{vougioukas2019video} and (iii) Lip2Wav~\cite{Prajwal_2020_CVPR}. Note that since we train using the same settings as Lip2Wav~\cite{Prajwal_2020_CVPR} for TCD-TIMIT dataset, we report the paper scores for all the comparison methods. Similarly, we take the paper scores from~\cite{vougioukas2019video} for GRID dataset (speaker-independent training settings).

\subsubsection{Metrics}
We evaluate our model using the standard speech metrics: Perceptual Evaluation of Speech Quality (PESQ) and short-time objective intelligibility measure (STOI). PESQ measures the overall perceptual quality of speech and STOI correlates with the intelligibility of speech. Also, to specifically evaluate the voice quality of the generated samples, we measure the distance ($L_1$) between the speaker embeddings of the generated and the ground-truth samples (termed speaker embedding distance (SED)).

\subsubsection{Results}
Table~\ref{table:grid_timit} shows the results of different models on GRID and TCD-TIMIT datasets. We see that our approach designed specifically for the unconstrained scenario performs slightly better or comparable to other methods when used in constrained settings. Also, we can observe that in terms of voice quality (SED metric), our method beats the existing approaches, thus indicating that we are able to preserve the voice of the identity to a large extent.

\begin{table}[ht]
    \centering
    \small
    \setlength{\tabcolsep}{3pt}
    \caption{Quantitative results on the constrained GRID~\cite{cooke2006audio} and TCD-TIMIT~\cite{harte2015tcd} datasets.}
    \vspace{-10pt}
    \begin{tabular}{l|ccc||ccc}
    \hline
        
    \textbf{Dataset} & \multicolumn{3}{c||}{GRID~\cite{cooke2006audio}} & \multicolumn{3}{c}{TCD-TIMIT~\cite{harte2015tcd}}  \\
    \hline
    \textbf{Method} & PESQ$\uparrow$ & STOI$\uparrow$ & SED$\downarrow$ & PESQ$\uparrow$ & STOI$\uparrow$ & SED$\downarrow$ \\
    \hline
    
    Imp. Vid2Speech~\cite{Ephrat2017ImprovedSR} & n/a & n/a & n/a & 1.23 & 0.49 & n/a\\
    
    GAN-based~\cite{vougioukas2019video} & 1.24 & 0.44 & n/a & 1.22 & 0.32 & n/a\\
    
    Lip2Wav~\cite{Prajwal_2020_CVPR} & 1.20 & 0.38 & 4.38 & 1.35 & \textbf{0.56} & 4.64 \\
    
    \hline
    
    \textbf{Ours} & \textbf{1.28} & \textbf{0.45} & \textbf{3.76} & \textbf{1.35} & 0.55 & \textbf{4.36}\\
    
    \hline
    
    \end{tabular}
    \vspace{-10pt}
    \label{table:grid_timit}
\end{table}

\begin{table*}[ht]
    \centering
    \small
    \setlength{\tabcolsep}{3.5pt}
    \caption{All models are pre-trained on LRW dataset and then trained on LRS2. We can see that we outperform all the competitive methods, especially on the challenging LRS2 data, which contains unseen speakers, words, poses and a large vocabulary.}
    
    \vspace{-10pt}
    \begin{tabular}{l|ccccccc||ccccccc}
    \hline
 
    \textbf{Dataset} & \multicolumn{7}{c||}{LRW~\cite{chung2016lip}} & \multicolumn{7}{c}{LRS2~\cite{chung2017lip}} \\
    \hline
    
    \textbf{Method} & PESQ$\uparrow$ & STOI$\uparrow$ & SED$\downarrow$ & FDSD$\downarrow$ & KDSD$\downarrow$ & LSE-C$\uparrow$ & LSE-D$\downarrow$ &
    PESQ$\uparrow$ & STOI$\uparrow$ & SED$\downarrow$ & FDSD$\downarrow$ & KDSD$\downarrow$ & LSE-C$\uparrow$ & LSE-D$\downarrow$\\
    
    \hline
    
    Imp. Vid2Speech~\cite{Ephrat2017ImprovedSR} & 0.65 & 0.09 & 6.01 & 5.645 & 10.2 & 1.782 & 10.43 & 0.59 & 0.30 & 6.25 & 4.275 & 3.1 & 2.009 & 8.424 \\
    
    GAN-based~\cite{vougioukas2019video} & 0.72 & 0.10 & 5.90 & 5.189 & 9.1 & 1.983 & 9.426 & 0.80 & 0.40 & 6.13 & 3.626 & 1.8 & 2.503 & 8.489\\
    
    Lip2Wav~\cite{Prajwal_2020_CVPR} & 1.19 & 0.54 & 5.73 & 1.831 & 1.1 & 2.526 & 8.286 & 0.58 & 0.28 & 6.22 & 10.71 & 15.5 & 1.874 & 11.48\\

    Seq2seq baseline & 1.01 & 0.50 & 6.16 & 4.306 & 7.7 & 2.396 & 8.412 & 0.97 & 0.43 & 6.47 & 3.840 & 2.8 & 1.991 & 8.532 \\
    
    Non seq2seq baseline & 1.05 & 0.49 & 6.17 & 4.112 & 7.1 & 2.282 & 8.441 & 0.96 & 0.44 & 6.43 & 3.803 & 2.3 & 2.078 & 8.536\\
    
    Ours w/o Content Encoder & 0.53 & 0.14 & 5.89 & 2.941 & 3.4 & 2.531 & 8.205 & 0.45 & 0.32 & 6.02 & 2.856 & 1.2 & 2.385 & 8.230 \\
    
    \hline
    \hline
    
    Lip-to-text~\cite{SubWordLR_arxiv_2022} + TTS~\cite{sv2tts} & 0.60 & 0.09 & 5.91 & 1.056 & 0.5 & 2.181 & 14.160 & 0.48 & 0.11 & 6.17 & 0.984 & 0.1 & 2.024 & 19.012 \\
    
    \hline
    \hline
    
    \textbf{Ours} & 0.78 & 0.15 & \textbf{5.65} & \textbf{1.638} & \textbf{0.8} & \textbf{2.538} & \textbf{8.173} & 0.60 & 0.34 & \textbf{5.95} & \textbf{1.273} & \textbf{0.2} & \textbf{2.507} & \textbf{8.155}\\

    \hline
    
    \end{tabular}
    \vspace{-10pt}
    \label{table:lrw_lrs2}
\end{table*}

\subsection{Evaluation in Unconstrained Settings}

\subsubsection{Baselines}
As no prior works in the multi-speaker lip-to-speech synthesis train on such unconstrained datasets, we extend previous models~\cite{Ephrat2017ImprovedSR, vougioukas2019video, Prajwal_2020_CVPR} with the same speaker embedding we use in our model and train all of them on the same dataset as ours. On the LRW dataset, we evaluate the publicly released multi-speaker Lip2Wav model. Additionally, to highlight the importance of our novel modules and facilitate more direct comparison, we implement the following baselines: (i) a non sequence-to-sequence encoder-decoder architecture, (ii) a sequence-to-sequence model with only L$_1$ reconstruction loss and without the VAE-GAN setup, (iii) A standard speech encoder trained from scratch instead of our pre-trained content encoder and (iv) Lip-to-text~\cite{SubWordLR_arxiv_2022} followed by text-to-speech (TTS)~\cite{sv2tts} model. 

\subsubsection{Metrics}
Explicitly modeling the stochastic nature of the problem is one of the major contributions of our work. Naturally, this allows our model to generate speech samples, which can differ from the original ground-truth. Thus, along with the standard speech evaluation metrics (PESQ, STOI) and our voice quality metric (SED), that directly evaluate the generated speech against a fixed ground-truth, we also evaluate our model using the perceptual metrics. Specifically, following the recent GAN-based TTS systems~\cite{main_metrics}, we propose to use: \textit{Frechet DeepSpeech Distance (FDSD)} and \textit{Kernel DeepSpeech Distance (KDSD)}, to evaluate the perceptual quality and the linguistic aspect of the generated speech. Note that we multiply KDSD scores with $10^3$ for better readability. Further, we also evaluate whether the output speech matches the lip movements using LSE-C (measures the confidence of lip-syncing) and LSE-D (measures an embedding level distance between the speech and lip-movements) metrics of~\cite{wav2lip}. We use the public implementations of these metrics for reliable comparison and reproducibility.

\subsubsection{Results}
Table~\ref{table:lrw_lrs2} compares our model with the different methods on the LRW and LRS2 datasets. 
We outperform existing approaches~\cite{Ephrat2017ImprovedSR, vougioukas2019video} and the baseline methods by a significant margin in perceptual metrics. Thus, although we under-perform in standard speech metrics (PESQ and STOI), we argue that our method is superior because perceptual metrics are more correlated to human judgement of intelligibility and speech quality. We further support this fact by providing qualitative results in the demo video on our website and conducting a human evaluation (Table~\ref{table:human_evals}). The standard metrics PESQ and STOI enforce one-to-one mapping, and thus are not ideal for evaluating our method. Also, GRID and TIMIT are constrained datasets with very less variations. On the other hand, LRW and LRS2 are more challenging and unconstrained datasets and our method is more effective on such challenging data. On the LRS2 dataset, where single-speaker methods such as Lip2Wav~\cite{Prajwal_2020_CVPR} fail to learn the audio-visual alignment, we achieve state-of-the-art perceptual metric scores. We encourage the reader to view the demo video for qualitative comparisons demonstrating the superiority of our approach. \\

\noindent
\textbf{Lip-to-text + TTS baseline:}
An additional baseline would be to use a state-of-the-art lip-to-text model~\cite{SubWordLR_arxiv_2022} and convert the predicted text transcripts to speech using a multi-speaker TTS model~\cite{sv2tts}. We can deduce the following from the scores reported in Table~\ref{table:lrw_lrs2}. The lip-to-text model trained on text transcripts is naturally far more accurate in predicting the word tokens than any lip-to-speech model. Thus, it achieves the best results in terms of intelligibility and perceptual quality metrics such as FDSD, KDSD. The fact that our lip-to-speech model comes close to the lip-to-text baseline for the same metrics shows that our approach captures the speech content most accurately. 

For other metrics like LSE-D that measures if the generated speech is in-sync with the video, we see that the output of lip-to-text baseline is not in-sync with the lip movements. The same content can be uttered in different ways (speeds, accent, prosody, voice), and the lip-to-text + TTS baseline cannot capture this. All the lip-to-speech models inherently achieve this to different extents and is an essential condition for the lip-to-speech task. Thus, ours is the best approach for the task of lip-to-speech synthesis.\\

\begin{figure}[t]
  \includegraphics[width=\linewidth]{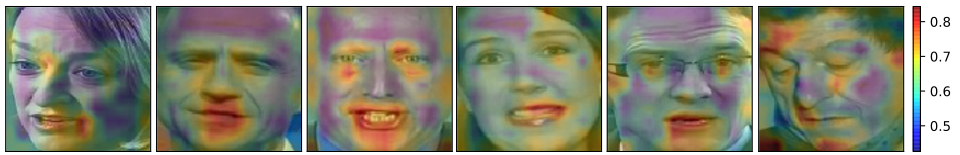}
  \vspace{-20pt}
  \caption{Activation maps of the visual encoder. Our model strongly attends to the lip region while generating speech, despite variations in head poses and the lip location.}
  \label{fig:actvation}
  \vspace{-10pt}
\end{figure}

\noindent
\textbf{Qualitative results:} In Figure~\ref{fig:actvation}, we plot the activation maps from the visual encoder to highlight that the model predominantly attends to the lip region. We encourage the reader to check our supplementary and demo video on our website for qualitative samples.\\

\noindent
\textbf{Computation cost:} We train our network using $4$ NVIDIA $2080$ Ti GPUs. The network consists of $18$M parameters and takes $0.5$-seconds to generate $1$-second of speech.

\subsection{Human Evaluations}

We perform human evaluations with the help of $20$ participants. The participant group spans members of $22 - 40$ years with an almost equal male-female ratio. We choose $15$ random samples from the LRS2 dataset~\cite{chung2017lip} and generate the results for all the comparison models. Participants rate the speech segments on a scale of $1 - 5$ based on: (A) Intelligibility (is the speech meaningful?) (B) Perceptual Quality (C) Sync Accuracy (is the generated speech in-sync with lip movements?) and (D) Voice Match. Table~\ref{table:human_evals} summarizes the mean scores of all the participants. Inline with the quantitative evaluations, the speech generated by our approach is of considerably higher quality and is more legible and natural. We also perform a Student’s T-Test for Table~\ref{table:human_evals} and compute the p-value to be $\approx 0.035$, indicating that the differences are statistically significant.

\begin{table}[ht]
    \centering
    \setlength{\tabcolsep}{8pt}
    \caption{(A) Intelligibility (is the speech meaningful?), (B) Perceptual Quality, (C) Sync Accuracy, (D) Voice Match. Our approach outputs meaningful, intelligible speech that matches lip movements and voice of the target person.}
    
    \vspace{-10pt}
    \begin{tabular}{lcccc}
    \hline

    Method & (A) & (B) & (C) & (D) \\ \hline 
    
    Imp. Vid2Speech~\cite{Ephrat2017ImprovedSR} & 2.02 & 1.98 & 1.74 & 1.13 \\
    WGAN-based~\cite{vougioukas2019video} & 2.17 & 2.43 & 2.19 & 2.01\\
    Lip2Wav~\cite{Prajwal_2020_CVPR} & 1.07 & 1.02 & 1.25 & 1.03\\
    Seq2seq baseline & 1.98 & 2.10 & 1.86 & 1.83 \\
    Non seq2seq baseline & 2.01 & 2.23 & 1.92 & 1.84 \\
    
    Ours w/o Content Encoder & 2.51 & 2.62 & 2.01 & 1.76 \\
    
    \hline
    
    \textbf{Ours} & \textbf{3.22} & \textbf{2.98} & \textbf{2.28} & \textbf{2.69}\\
    
    \hline
    
    \end{tabular}
    \vspace{-10pt}
    \label{table:human_evals}
\end{table}

\begin{figure}[t]
  \includegraphics[width=\linewidth]{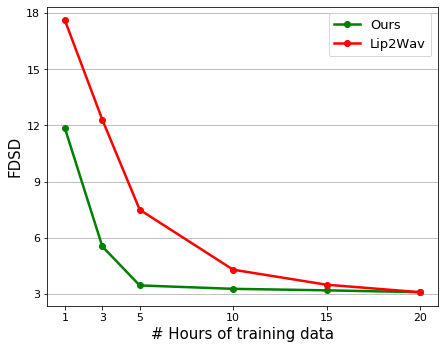}
  \vspace{-20pt}
  \caption{Fine-tuning our pre-trained multi-speaker model consistently outperforms the current best single-speaker model (FDSD lower is better) in the low data regime.}
  \label{fig:single_speaker}
  \vspace{-10pt}
\end{figure}

\subsection{Adapting to Single-Speaker Lip-to-Speech}

Our model can generate speech for arbitrary speakers, which is highly beneficial for applications where there is almost no training data available for that speaker. However, in a few applications, it is possible to obtain some data of a target speaker for fine-tuning. Current single-speaker models need nearly $20$ hours of data to produce impressive results. This is really difficult to obtain in many scenarios. Our multi-speaker model can resolve this issue to a large extent - we can fine-tune our pre-trained multi-speaker model on a small amount of speaker-specific data and achieve impressive personalized results. By using only $25\%$ of the training data ($5$ hours), we can nearly match the single-speaker model's performance trained with $20$ hours.

We fine-tune our network on the speakers in the Lip2Wav dataset~\cite{Prajwal_2020_CVPR}. We vary the number of hours in the train set and train the current single-speaker state-of-the-art Lip2Wav model~\cite{Prajwal_2020_CVPR} and fine-tune our multi-speaker model. We plot the variation of the FDSD metric with the training data size in Figure~\ref{fig:single_speaker}. We can clearly see that in the low data regime, pre-training on multi-speaker data vastly outperforms the best single-speaker model trained from scratch.

\section{Ablation Studies}
\label{section:ablation}    

We perform ablations to assess the effect of our key design choices. The results are on the unseen LRS2 test set.
 
\subsection{Impact of Each Discriminator}

We use two discriminators in our final model, one for enforcing better voice and style attributes and another for enforcing realistic speech. We assess the importance of using each of them in Table~\ref{table:discs}. We can see that despite getting a minor improvement in lip-sync metrics, both the discriminators enforce better overall speech generation as observed by the speech metrics.

\vspace{-5pt}
\begin{table}[ht]
    \centering
    \setlength{\tabcolsep}{5pt}
    \caption{The discriminators enforce our model to produce meaningful and realistic speech outputs.}
    
    \vspace{-10pt}
    \begin{tabular}{lcccc}
    \hline

    Method & FDSD$\downarrow$ & KDSD$\downarrow$ & LSE-C$\uparrow$ & LSE-D$\downarrow$\\ 
    \hline 
    Ours w/o both Discs & 4.055 & 2.9 & 2.188 & 8.199 \\
    Ours w/o WGAN & 3.916 & 2.7 & 2.294 & 8.194 \\
    Ours w/o Voice Disc & 4.310 & 3.6 & 2.319 & 8.189 \\
    \hline
    \textbf{Ours} & \textbf{1.273} & \textbf{0.2} & \textbf{2.507} & \textbf{8.155} \\
    \hline
    
    \end{tabular}
    \vspace{-5pt}
    \label{table:discs}
\end{table}

\subsection{Importance of Local and Global Alignment}
Table~\ref{table:global_vs_local} shows that optimizing both local and global KL-divergence together improves the alignment between lip and content distributions, thus improving the overall performance. Training with either of the losses in isolation leads to inferior results.

\begin{table}[ht]
    \centering
    \setlength{\tabcolsep}{3pt}
    \caption{Optimizing both the global and local KL-divergence loss improves the overall quality of the results.}
    
    \vspace{-10pt}
    \begin{tabular}{lcccc}
    \hline

    Method & FDSD$\downarrow$ & KDSD$\downarrow$ & LSE-C$\uparrow$ & LSE-D$\downarrow$ \\ 
    \hline 

    Ours w/o local KL div &  2.883 & 3.2 & 2.340 & 8.249 \\
    Ours w/o global KL div & 5.040 & 6.8 & 2.003 & 8.937 \\
    \hline
    \textbf{Ours} & \textbf{1.273} & \textbf{0.2} & \textbf{2.507} & \textbf{8.155} \\
    \hline
    
    \end{tabular}
    \vspace{-10pt}
    \label{table:global_vs_local}
\end{table}

\section{Limitations and Future Directions}
\label{section:limitations}

Unconstrained lip-to-speech synthesis is far from a solved problem - there is still a considerable room for improvement. Ours is the first attempt to design a model which can generate speech for any speaker in any voice. We specifically deal with the issue of learning speech-lip correspondences and handling the homopheme ambiguities. However, there are still multiple unresolved problems. For example, our model struggles when there is a drastic movement of the head while speaking and if the head is non-frontal. Another issue is that we can get output sounds that do not form the right words or phrases. This is because it is hard to learn a language model in the speech modality compared to lip-to-text models that train on text transcripts. We hope our efforts in this work lead to new future directions that tackle some of the aforementioned issues.

\section{Conclusion}
\label{section:conclusion}
In this work, we address the problem of the unconstrained lip-to-speech synthesis for the first time. We extensively discuss the challenges this problem presents, due to which the existing approaches fail to scale to such unconstrained settings. To tackle these challenges, we propose a VAE-GAN model trained to explicitly handle the stochastic nature of the task. Our approach produces significantly more intelligible, realistic speech outputs compared to all other models. We justify the use of different parts of our architecture with numerous ablation studies. We believe that our core idea of handling stochasticity can encourage future efforts to enable advanced versions of lip-to-speech and lip-to-text generation systems for arbitrary languages and speakers in the wild.

\bibliographystyle{ACM-Reference-Format}
\balance
\bibliography{sample-base}

\newpage
\appendix

\section{SUPPLEMENTARY MATERIAL}

\subsection{Additional Experiments and Ablation Studies}

In this section, we report additional experiments and ablation studies to get further insights and better understand the different aspects of our network. The experimental setup is identical to that of the ablation studies in the main paper.

We also encourage the reader to view the demo video containing the results and comparisons.

\subsubsection{Comparisons on LRS3~\cite{afouras2018lrs3} Dataset}
\label{comparison_methods}
We compare our work on the LRS3~\cite{afouras2018lrs3} test split. LRS3 dataset was collected from TedX videos and consists of a large vocabulary and majorly profile face videos. Further, the videos also have very different lighting conditions and extreme head-motions when compared to LRS2~\cite{chung2017lip} dataset. Please note that, we do not fine tune our model on LRS3 dataset; thus evaluating on it highlights our model's generalization ability and robustness on completely unseen speakers. As seen in Table~\ref{table:lrs3}, our method performs well on LRS3 videos containing unseen vocabulary, identities, voices, and profile views, clearly indicating the robustness of our approach. 

\begin{table}[ht]
    \centering
    \setlength{\tabcolsep}{3pt}
    \caption{Quantitative comparison on LRS3 dataset~\cite{afouras2018lrs3}. We can see that we outperform all the competitive methods, even in a very different setting in LRS3, which contains unseen speakers, words, and a large number of profile views. Note that our model is not fine-tuned on the LRS3 dataset.}
    \vspace{-10pt}
    \begin{tabular}{l|cccc}
    \hline
 
    \textbf{Dataset} & \multicolumn{4}{c}{LRS3~\cite{afouras2018lrs3}} \\
    \hline
    \textbf{Method} & FDSD$\downarrow$ & KDSD$\downarrow$ & LSE-C$\uparrow$ & LSE-D$\downarrow$\\
    \hline
    
    Imp. Vid2Speech~\cite{Ephrat2017ImprovedSR} & 5.286 & 5.4 & 1.929 & 8.435 \\
    GAN-based~\cite{vougioukas2019video} & 4.589 & 4.2 & 2.031 & 9.372 \\
    Lip2Wav~\cite{Prajwal_2020_CVPR} & 12.663 & 16.2 & 1.632 & 11.465\\
    Seq2seq baseline & 4.821 & 4.8 & 1.880 & 8.568 \\
    Non seq2seq baseline & 4.766 & 4.2 & 1.874 & 8.579 \\
    Ours w/o Content Encoder & 4.054 & 2.9 & 2.041 & 8.312 \\
    \hline
    \textbf{Ours} & \textbf{3.148} & \textbf{1.8} & \textbf{2.063} & \textbf{8.256} \\
    \hline
    
    \end{tabular}
    \label{table:lrs3}
\end{table}

\subsubsection{Near Frontal vs. Non-frontal Videos}
We study the extent to which the performance deteriorates if the face view moves towards a non-frontal profile view. As expected and also observed in past lip-reading works~\cite{Afouras2018DeepLR}, our model also has room for improvement when handling non-frontal talking faces. 

\begin{table}[ht]
    \centering
    \setlength{\tabcolsep}{4pt}
    \caption{Similar to other lip-reading models~\cite{Afouras2018DeepLR}, our model also has room for improvement when dealing with non-frontal views.}
    \vspace{-10pt}
    \begin{tabular}{lcccc}
    \hline

    Method & LSE-C$\uparrow$ & LSE-D$\downarrow$ & FDSD$\downarrow$ & KDSD$\downarrow$ \\ 
    \hline 
    
    Near frontal & 2.608 & 8.016 & 1.351 & 0.3 \\
    Non-frontal & 2.491 & 8.058 & 3.473 & 1.0\\
    \hline
    
    \end{tabular}
    \label{table:frontal}
\end{table}

\subsubsection{What Kind of Visual Input is the best?}
We compare different forms of visual inputs, such as feeding only the lower-half of the face and pre-trained face embeddings~\cite{facenet}. Providing the full face crop performs the best, as shown in Table~\ref{table:visual_inp} and also reflected by activation maps near the eye regions in Figure 4 of the main paper.

\begin{table}[H]
    \centering
    \setlength{\tabcolsep}{4pt}
    \caption{Feeding the full face crop produces the best results.}
    \vspace{-10pt}
    \begin{tabular}{lcccc}
    \hline

    Method & LSE-C$\uparrow$ & LSE-D$\downarrow$ & FDSD$\downarrow$ & KDSD$\downarrow$ \\ 
    \hline 
    Facenet emb~\cite{facenet} & 1.931 & \textbf{7.664} & 8.641 & 8.7\\
    Lower half (ours) & 2.338 & 8.173 & 3.224 & 2.8\\
    \textbf{Full face (ours)} & \textbf{2.507} & 8.155 & \textbf{1.273} & \textbf{0.2}\\
    \hline
    
    \end{tabular}
    \label{table:visual_inp}
\end{table}

\subsubsection{Sampling strategy of VAE at train-time}

In 3.2.4 of the main paper, we mention that there are three ways of sampling the points to decode from during the training: (i) only from the lip distribution, (ii) only from the speech content distribution, (iii) alternately sample from both the distributions. We mentioned that we obtain good convergence and intelligible results only by following (ii), because it allows the network to learn excellent latent representations of the speech, which can help overall learning. We experimentally verify this in Table~\ref{table:vae}, by showing that sampling from other distributions, i.e. (i) Lip distributions and (iii) Alternately sampling from both the distributions, leads to far worse results.

\begin{table}[ht]
    \centering
    \setlength{\tabcolsep}{3pt}
    \caption{Sampling solely from the speech distribution during training enables the decoder to learn to generate realistic, accurate outputs.}
    \vspace{-10pt}
    \begin{tabular}{lcccc}
    \hline

    Method & LSE-C$\uparrow$ & LSE-D$\downarrow$ & FDSD$\downarrow$ & KDSD$\downarrow$ \\ 
    \hline 
    
    Lip dist. & 2.261 & 8.195 & 4.036 & 2.7\\
    Speech content dist. & \textbf{2.507} & \textbf{8.155} & \textbf{1.273} & \textbf{0.2}\\
    Alternate sampling & 2.241 & 8.320 & 3.487 & 2.4\\
    
    \hline
    
    \end{tabular}
    \label{table:vae}
\end{table}

\subsubsection{Auto-encoder vs. VAE}
We assess the importance of mapping the lip and the content distributions using a VAE. In Table~\ref{table:vae}, we show that removing the variational aspect and using just a naive auto-encoder approach results in poor speech generation. Interestingly, while the network learns comparable audio-visual correspondence, the speech is not intelligible or meaningful, as indicated by the speech metrics.

\begin{table}[ht]
    \centering
    \setlength{\tabcolsep}{6pt}
    \caption{Using a VAE enables the model to generate meaningful, high-fidelity speech outputs.}
    
    \vspace{-10pt}
    \begin{tabular}{lcccc}
    \hline

    Method & FDSD$\downarrow$ & KDSD$\downarrow$ & LSE-C$\uparrow$ & LSE-D$\downarrow$ \\ 
    \hline 
    
    Auto-encoder & 6.015 & 5.3 & 2.002 & 8.431 \\
    \textbf{Ours (VAE)} & \textbf{1.273} & \textbf{0.2} & \textbf{2.507} & \textbf{8.155} \\
    \hline
    
    \end{tabular}
    \label{table:vae}
\end{table}

\subsubsection{Model's variation across speaker attributes}

In Table~\ref{table:gender}, we evaluate the performance of our model across gender of the identities. We automatically classify the LRS2 test set into male and female speakers using a gender detection tool~\cite{ar2018cvlib}. From the table, we can clearly observe that there is no significant variation in performance across gender of the identities.

\begin{table}[ht]
    \centering
    \setlength{\tabcolsep}{5pt}
    \caption{There is no distinctive variation of performance across gender of the speakers.}
    \vspace{-10pt}
    \begin{tabular}{lcccc}
    \hline

    Gender & LSE-C$\uparrow$ & LSE-D$\downarrow$ & FDSD$\downarrow$ & KDSD$\downarrow$ \\ 
    \hline 
    
    Female & 2.549 & 8.138 & 1.633 & 0.8 \\
    Male & 2.424 & 8.233 & 1.703 & 0.8 \\
    
    \hline
    
    \end{tabular}
    \label{table:gender}
\end{table}

\balance

\begin{figure}[H]
    \centering
    \includegraphics[width=\linewidth]{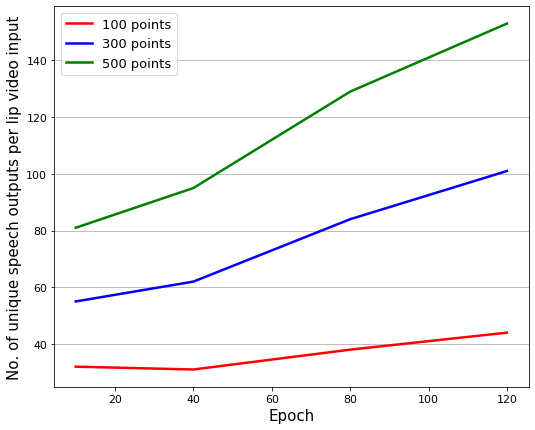}
    \vspace{-20pt}
    \caption{We plot the average number of unique speech outputs generated by our model for every input lip video. We show this "generative strength"~\cite{creativity} at different stages of training, and we can see that our model captures more variations in the latent space as the training progresses.}
    \label{fig:generative_strength}
\end{figure}

\subsection{Generative Strength of our Lip-to-Speech Model}

Lip to speech synthesis is a highly ambiguous task, and many speech outputs are possible for the same input lip sequence. For instance, the variations in voice, speech amplitude, intonation, prosody, and emotion do not clearly correlate with the lips. Moreover, the content to generate is also ambiguous due to the presence of homophenes. Our VAE-GAN model is the first architecture that is capable of modeling these variations in the latent space. In fact, for the same input lip sequence, we can generate different output speech sequences, by sampling different points from the lip distribution. 

Evaluating the generative capabilities of a VAE/GAN has been explored in a previous work~\cite{creativity}, and we adopt the same metric here. The "Generative Strength" metric determines the average percentage of unique speech samples generated for every input lip sequence. To compute this metric, we sample $N$ points ($N = {100, 300, 500}$) for each input lip video from the test set of LRS2. Hence, we obtain $N$ speech outputs for every LRS2 test video. We consider a generated spectrogram output as ``unique", if its L2 distance from the remaining $N - 1$ spectrograms is at least $\delta=0.5$. Note that this is a high enough L2 threshold, as the generated audio samples are clearly distinct to hear. In Figure~\ref{fig:generative_strength}, we plot the average count of unique speech outputs per input lip video at different stages of the model training. 
We can see that our model captures more variations for the same input lip sequence as the training progresses.

\subsection{Lip2Wav fails to learn the attention alignment}
Lip2Wav~\cite{Prajwal_2020_CVPR} is a sequence-to-sequence model with attention, and was proposed for the \textit{single-speaker} lip-to-speech task. The attention mechanism allows the decoder to look at the correct frame's lip movements while decoding. The model learns diagonal attention~\cite{Prajwal_2020_CVPR} upon convergence.

\begin{figure}[H]
    \centering
    \includegraphics[width=\linewidth]{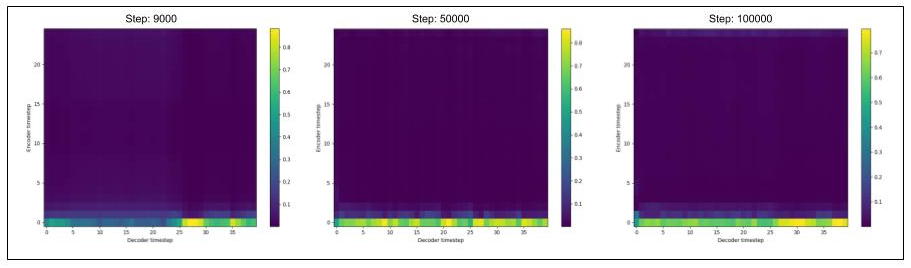}
    \caption{Attention alignment plots at various training stages indicating that Lip2Wav~\cite{Prajwal_2020_CVPR} fails to learn temporal attention in unconstrained multi-speaker setting.}
    \label{fig:lip2wav_plot}
\end{figure}

We observed that when training on challenging sentence-level, multi-speaker datasets such as LRS2~\cite{chung2017lip} with a vast number of voices and vocabulary, it fails to learn the temporal attention alignment. We mention this in the main paper, but we also add the final alignment plots of the trained model in Fig~\ref{fig:lip2wav_plot} to clearly show the failure of this model to learn audio-visual correspondence in such unconstrained settings. The public code\footnote{github.com/Rudrabha/Lip2Wav} provided by the authors is used for this implementation.

\end{document}